\documentclass[journal]{IEEEtran}

\usepackage{inputenc}
\usepackage{mathtools}
\usepackage{color}
\usepackage{array}
\usepackage{verbatim}
\usepackage{float}
\usepackage{amsmath}
\usepackage{amsthm}
\usepackage{amssymb}
\usepackage{graphicx}
\usepackage{longtable}
\usepackage{multirow}
\usepackage{balance}
\usepackage{mathtools}
\usepackage{tikz}
\usepackage{pgf-pie}

\usepackage[unicode=true,
bookmarks=false,
breaklinks=false,pdfborder={0 0 1},colorlinks=false]
{hyperref}
\hypersetup{
	colorlinks,bookmarksopen,bookmarksnumbered,citecolor=red,urlcolor=red}
\usepackage{cite}
\usepackage{nomencl} 
\makenomenclature
\usepackage{longtable}

\usepackage{indentfirst}
\usepackage{textcomp}
\usepackage[T1]{fontenc}
\usepackage{amsthm}
\usepackage{amsmath,amstext,amsfonts}
\usepackage{amssymb}
\usepackage{mathtools}
\usepackage[cmintegrals]{newtxmath}
\usepackage{adjustbox}
\usepackage{caption}
\usepackage{bm}  
\usepackage{siunitx}
\usepackage{physics}

\usepackage{graphicx}
\graphicspath{%
	{Figures/}
	{Figures/TiKz/}
	{Figures/Ipe/}
	{Figures/XFig/}
	{Figures/Inkscape/}
	{Figures/Matlab/}
	{Figures/Scanned/}
	{Figures/Saved/}
}
\usepackage{tikz}
\usepackage[list=true,labelformat=simple]{subcaption}
\usepackage{cite}
\usepackage{url}
\usepackage{xcolor}
\usepackage{siunitx}

\usepackage[inline]{enumitem}

\usepackage{algorithm,algpseudocode,float}
\usepackage{setspace} 

\algnewcommand{\algorithmicand}{\textbf{ and }}
\algnewcommand{\algorithmicor}{\textbf{ or }}
\algnewcommand{\AlgAnd}{\algorithmicand}
\algnewcommand{\AlgOr}{\algorithmicor}

\usepackage[noabbrev]{cleveref}  
\Crefname{figure}{Fig.}{Figs.}

\usepackage{tabularx,booktabs}
\newcolumntype{C}{>{\centering\arraybackslash}X} 
\usepackage{lipsum}

\makeatletter
\let\oldforeign@language\foreign@language
\DeclareRobustCommand{\foreign@language}[1]{%
	\lowercase{\oldforeign@language{#1}}}

\floatstyle{ruled}
\newfloat{algorithm}{tbp}{loa}
\providecommand{\algorithmname}{Algorithm}
\floatname{algorithm}{\protect\algorithmname}

\let\oldforeign@language\foreign@language
\DeclareRobustCommand{\foreign@language}[1]{%
	\lowercase{\oldforeign@language{#1}}}

\ifCLASSINFOpdf
\else
\fi

\ifCLASSINFOpdf
\else
\fi

\def\ps@IEEEtitlepagestyle{%
	\def\@oddhead{\parbox[t][\height][t]{\textwidth}{\centering \scriptsize
			Personal use of this material is permitted. Permission from the author(s) and/or copyright holder(s), must be obtained for all other uses. Please contact us and provide details if you believe this document breaches copyrights.\\
			\noindent\makebox[\linewidth]{}
		}\hfil\hbox{}}%
	\def\@evenhead{\scriptsize\thepage \hfil \leftmark\mbox{}}%
	\def\@oddfoot{\parbox[t][\height][l]{\textwidth}{
			\vspace{-20pt}{\rule{\textwidth}{0.4pt}}\\ \footnotesize			{\bf{\footnotesize\textcolor{red}{A. M. Ali, L. Tirel, and H. A. Hashim, "Novel Multi-Agent Action Masked Deep Reinforcement Learning for General Industrial Assembly Lines Balancing Problems," Journal of Automation and Intelligence, vol. 1, pp. 100096, 2025.}}} doi: \href{https://doi.org/10.1016/j.jai.2025.07.001}{10.1016/j.jai.2025.07.001}\\
			\noindent\makebox[\linewidth]
		}\hfil\hbox{}}%
	\def\@evenfoot{\MYfooter}}

\makeatother
\begin{document}
	\bstctlcite{IEEEexample:BSTcontrol}

\title{Novel Multi-Agent Action Masked Deep Reinforcement Learning for General Industrial Assembly Lines Balancing Problems}

\author{Ali Mohamed Ali, Luca Tirel, and Hashim A. Hashim
	\thanks{This work was supported in part by the National Sciences and Engineering Research Council of Canada (NSERC), under the grants RGPIN-2022-04937 and DGECR-2022-00103.}
	\thanks{A. M. Ali, L. Tirel, and H. A. Hashim are with the Department of Mechanical
		and Aerospace Engineering, Carleton University, Ottawa, ON, K1S-5B6,
		Canada (e-mail: hhashim@carleton.ca)}
}



\maketitle
\begin{abstract}
	Efficient planning of activities is essential for modern industrial assembly lines to uphold manufacturing standards, prevent project constraint violations, and achieve cost-effective operations. While exact solutions to such challenges can be obtained through Integer Programming (IP), the dependence of the search space on input parameters often makes IP computationally infeasible for large-scale scenarios. Heuristic methods, such as Genetic Algorithms, can also be applied, but they frequently produce suboptimal solutions in extensive cases. This paper introduces a novel mathematical model of a generic industrial assembly line formulated as a Markov Decision Process (MDP), without imposing assumptions on the type of assembly line a notable distinction from most existing models. The proposed model is employed to create a virtual environment for training Deep Reinforcement Learning (DRL) agents to optimize task and resource scheduling. To enhance the efficiency of agent training, the paper proposes two innovative tools. The first is an action-masking technique, which ensures the agent selects only feasible actions, thereby reducing training time. The second is a multi-agent approach, where each workstation is managed by an individual agent, as a result, the state and action spaces were reduced. A centralized training framework with decentralized execution is adopted, offering a scalable learning architecture for optimizing industrial assembly lines. This framework allows the agents to learn offline and subsequently provide real-time solutions during operations by leveraging a neural network that maps the current factory state to the optimal action. The effectiveness of the proposed scheme is validated through numerical simulations, demonstrating significantly faster convergence to the optimal solution compared to a comparable model-based approach.
\end{abstract}


\section*{Nomenclature}
	\begin{tabular}{@{} l l @{}}
		$ i $ & Subscript indicating workstations \\
		$ j $ & Subscript indicating tasks \\
		$ r $ & Subscript indicating resources \\
		$ z $ & Subscript indicating a specific action in the actions space \\
		$ k $ & Discrete time instant \\
		$ I $ & Set of workstations \\
		$ J $ & Set of tasks \\
		$ R $ & Set of resources \\
		$ \delta_s $ & Dimension of the state \\
		$ \delta_A $ & Dimension of the centralized actions space \\
	\end{tabular}
	
	\begin{tabular}{@{} l l @{}}
		$ \delta_{A_{i}} $ & Dimension of the decentralized actions space \\
		$ \mathbf{O} $ & Workstations' capacities in terms of tasks \\
		$ \mathbf{U} $ & Workstations' capacities in terms of resources \\
		$ \mathbf{D} $ & Tasks' durations workstation-dependent \\
		$ \mathbf{F} $ & Tasks' deadlines time instants \\
		$ \mathbf{P} $ & Tasks' time precedence constraints matrix \\
		$ \mathbf{C} $ & Tasks' required amount of resources \\
		$ \mathbf{G} $ & Resources of each type available in the assembly line \\
		$ A $ & Centralized tasks actions space \\
		$ A_i $ & Decentralized tasks actions space \\
		$ A_s[k] $ & Subset of feasible actions at time $k$ \\
		$ \mathbf{a}[k] $ & Centralized control action for task assignment \\
		$ \mathbf{y}[k] $ & Centralized control action for resources assignment \\
		$ \mathbf{a}_i[k] $ & Decentralized control action for task assignment \\
		$ \mathbf{o}[k] $ & Workstations' states of occupancies \\
		$ \mathbf{e}[k] $ & Tasks' states of execution \\
		$ \mathbf{f}[k] $ & Tasks' states of finish \\
		$ \mathbf{d}[k] $ & Tasks' current remaining durations \\
		$ \mathbf{g}[k] $ & Resources levels in the assembly line \\
		$ \mathbf{r}[k] $ & Resources stored for the tasks executions \\
		$ \mathbf{b}[k] $ & Resources currently stored in a specific workstation \\
		$ \mathbf{m}[k] $ & Current centralized action mask \\
		$ \mathbf{m}_i[k] $ & Current decentralized action mask \\
		$ \mathbf{h}[k] $ & Ending state flag \\
		$ \mathbf{s}[k] $ & State of the factory at time $k$ \\
		$ k_{\text{end}} $ & Episodic ending time \\
		$ k_{\text{opt}} $ & Best achievable ending time computed via O.C. \\
		$ \mathcal{E} $ & Environment \\
		$ \mathcal{A}_i $ & Agent $i$-th \\
		$ \mathbf{p} $ & Action probabilities \\
		$ \mathbf{p_f} $ & Action probabilities filtered based on the mask \\
		$ f_{\text{learn}} $ & Learning frequency of the PPO \\
		$ \mathbf{s^F} $ & Fictitious state \\
		$ \mathbf{m^F} $ & Fictitious actions mask \\
	\end{tabular}

\section{Introduction}\label{sec1}

The growing complexity of activities in large, modern industries necessitates advanced planning systems for manufacturing and assembly lines to optimize key performance indicators while adhering to specific operational constraints. Assembly Line Balancing Problems (ALBPs) are well-established, challenging nonlinear programming problems characterized by context-specific formulations and significant computational demands \cite{sivasankaran2014literature}. During the execution phase, unforeseen disruptions such as equipment breakdowns can derail planned schedules, and the time-intensive process of recalculating solutions may render real-time adjustments impractical, leading to project delays and substantial unplanned costs. Effective assembly line balancing directly influences productivity, throughput, and cost-efficiency, making it a critical factor for industries striving to maintain a competitive edge \cite{oztemel2020literature}. In sectors with complex production processes, such as automotive and electronics manufacturing, ALBP plays a pivotal role in ensuring timely delivery, maintaining quality control, and responding efficiently to customer demands, thus driving profitability and operational excellence \cite{sivasankaran2014literature}. ALBP can be conceptualized as a control problem centered on the efficient assignment of tasks and resources to satisfy constraints while optimizing key performance metrics \cite{salveson1955assembly}. Artificial intelligence have been extensively applied to address complex engineering applications \cite{dong2024soft, tirel2024novel, jonnalagadda2024segnet, haddad2022deep, jonnalagadda2024comprehensive} and This conceptualization allows ALBP to be addressed through reinforcement learning (RL) approaches, where agents learn to allocate tasks and resources on the assembly line to maximize an objective function. Recent studies \cite{zhai2024explainable, dong2024soft, baltes2023deep, haddad2022deep, arishi2022machine} have successfully applied Deep Reinforcement Learning (DRL) frameworks to various allocation and control problems across fields such as blockchain, energy demand management, humanoid robotics, and traffic signal control. These advancements highlight DRL as a promising approach for effectively addressing the challenges associated with assembly line balancing.

%
%
\section{Related Work}
\label{literature_review}
\subsection{Optimal Control in ALBPs}
An assembly line consists of multiple workstations, each tasked with performing specific operations on items within a designated time interval, known as the cycle time. Solving the ALBP requires the optimal assignment of tasks to workstations while ensuring compliance with predefined constraints. Various formulations of ALBP are explored in \cite{battaia2013taxonomy}, which categorize the problem into two primary types: the Single Assembly Line Balancing Problem (SALBP) and the more comprehensive Generalized Assembly Line Balancing Problem (GALBP).

The work presented in \cite{salveson1955assembly} represents an early attempt to address ALBPs. Two primary families of ALBP problems were identified in \cite{salveson1955assembly}: the SALBP and the GALBP. The SALBP is characterized by single-element production, a serial line layout, a fixed common cycle time, a serial unilateral line, precedence constraints, and deterministic operation times. The GALBP formulation relaxes some of the assumptions of the SALBP to accommodate more realistic scenarios. Specifically, GALBPs allow for the consideration of multi-model processes, zone constraints, delays, parallel stations, and more complex layouts, among other factors. This broader class of problems includes the U-shaped Assembly Line Balancing Problem (UALBP) and the Mixed-Model Assembly Line Balancing Problem (MMALBP) \cite{kucukkoc2015balancing}. Another classification of ALBP is based on the objective function \cite{battaia2013taxonomy}. Type-1 focuses on minimizing the number of workstations required to meet a fixed finishing time while optimizing resource utilization and maintaining production rates. In contrast, Type-2 aims to minimize the finishing time for a given number of workstations. Exact solutions to the SALBP are typically derived by repeatedly solving instances of the closely related SALBP-1.
For instance, \cite{gyulai2017scheduling} proposes a two-level approach, where the first level estimates production lot sizes, while the second level focuses on rescheduling planned tasks. In \cite{koskinen2020rolling}, the authors developed an Integer Programming (IP) model to optimize operational planning for Printed Circuit Board (PCB) manufacturing. Historically, methods such as linear programming, integer programming, dynamic programming, and branch-and-bound have been employed to address ALBPs. However, these traditional techniques have increasingly been supplanted by search heuristics, such as genetic algorithms, which offer enhanced computational efficiency, as demonstrated in \cite{wang2021genetic, ccil2020mathematical, dalle2019designing, chen2023novel}. Recent studies have also focused on improving the robustness of these solutions. For example, \cite{gyulai2015robust} formulated a mixed-integer programming model to optimally allocate jobs and workers to workstations, leveraging historical data to reliably estimate optimal capacities. Additionally, \cite{manzini2018integrated} presented an integrated approach for designing reconfigurable assembly systems, accounting for potential future demand scenarios to mitigate reconfiguration costs.
%
%
\subsection{Deep Reinforcement Learning for ALBPs}
A key advantage of DRL methodologies is their adaptability. Once a model is trained to address a specific problem, solutions do not need to be recomputed in the event of unforeseen disruptions. Instead, agents can be retrained or fine-tuned for new scenarios, providing substantial flexibility \cite{kang2020machine}. However, a common challenge with Reinforcement Learning (RL) techniques is that they may yield solutions of lower quality compared to traditional methods, which has led to the development of hybrid approaches. For example, \cite{buddala2019two} introduced a hybrid approach that combines an RL-based scheduling algorithm with classical Optimal Control (OC). In \cite{zhou2020deep}, the authors proposed a smart scheduling agent based on a DRL algorithm, which demonstrated robustness in handling uncertainties such as unexpected task arrivals or equipment malfunctions. Similarly, \cite{wang2021dynamic} addressed an industrial scheduling problem under uncertain conditions by utilizing a DRL-based Proximal Policy Optimization (PPO) algorithm. This study will further explore advanced variants of both the on-policy PPO algorithm \cite{schulman2017proximal} and the off-policy Deep Q-Network (DQN) algorithm \cite{mnih2015human} to solve a Fully Observable Markov Decision Process (MDP). More recently, \cite{GEURTSEN2023170} examined optimal maintenance planning for scheduling in a serial production line, showcasing the applicability of DRL to complex industrial scheduling problems.
%
%
\subsection{Action Masking in Reinforcement Learning}
Classical optimal approaches define the solution space through constraints, with feasible solutions identified within this space. In contrast, in RL, satisfying constraints may lead to infeasible state transitions, requiring careful action selection to ensure feasibility. This can be addressed using a state-dependent function that maps the action space, distinguishing feasible actions from infeasible ones. One of the earliest efforts to extend the Markov Decision Process (MDP) with a feedback mechanism to reject invalid actions is presented in \cite{alshiekh2018safe}. This framework introduced "shielding" a technique that monitors an agent's actions and corrects them if they violate specified safety properties. In \cite{huang2020closer}, the authors explored action masking within policy gradient algorithms, defining and comparing various strategies for handling invalid actions and offering valuable theoretical insights. Notably, they demonstrated that a state-dependent action masker can function as a differentiable component, thereby enabling valid policy gradient updates. Building on this work, \cite{elsayed-aly21__safe_multiagent_reinforcement_learning_via_shielding} extended these principles to a multi-agent framework. They synthesized multiple shields based on a factorization of the joint state space across agents, with each shield monitoring a subset of agents, ensuring constraint adherence at every time step. Both action masking and shielding techniques have proven effective in eliminating unsafe and infeasible states during training. Action masking filters the action space to allow only feasible actions through a state-dependent function while shielding monitors the agent's actions and corrects them only when they violate predefined specifications. In the proposed work, action masking technique is employed to manage the numerous constraints inherent in the assembly line balancing problem. Without this approach, the large number of constraints could result in many futile actions, thereby increasing training time \cite{huang2020closer}.

\subsection{Multi Agent Reinforcement Learning}
The primary concept in Multi-Agent Reinforcement Learning (MARL) lies in decentralizing a single learning entity into multiple agents \cite{dai2022distributed}. While the overall complexity of MARL schemes can vary significantly depending on the domain, the core idea involves each agent learning its own policy so that, collectively, all agents achieve the system’s overarching goal. An extensive review of different types of MARL is provided in \cite{bucsoniu2010multi}, where the authors categorize tasks as cooperative, competitive, or general (neither purely cooperative nor competitive) and highlight a key challenge in this field: the formal articulation of a shared learning objective. Specifically, MARL algorithms must be designed to avoid complete independence from other agents or mere behavioral tracking, both of which could hinder convergence. A significant contribution to MARL theory was made in \cite{oroojlooy2023review}, where the authors developed a multi-agent Q-learning framework and proved convergence to a Nash equilibrium under certain conditions. Their findings demonstrated that convergence requires only that each action is attempted and each state is visited, enabling each agent to learn optimal Q-values without assuming specific behaviors of other agents, provided the immediate rewards of other agents are observable. In \cite{gupta2017cooperative}, the authors addressed cooperative multi-agent control using DRL, introducing parameter sharing among agents' policies. They extended algorithms such as DQN, Deep Deterministic Policy Gradient (DDPG), and Trust Region Policy Optimization (TRPO) to multi-agent settings, thereby ensuring scalability with continuous action spaces. Additional effective value-based approaches in MARL were proposed in \cite{rashid2020weighted} and \cite{wang2020qplex}. Practical recommendations for implementing MARL policy gradient algorithms, aimed at achieving optimal convergence, were provided in \cite{fu22__revisiting_some_common_practices_in_cooperative_multiagent_reinforcement_learning}. Furthermore, \cite{yu2022surprising} identified PPO as an effective algorithm for cooperative multi-agent games. In \cite{peng2021facmac}, the authors proposed a multi-agent centralized policy gradient method with non-monotonic factorization, capable of handling both discrete and continuous action spaces.

\subsection{Paper Contributions}
Motivated by the excessive computational burden of solving the ALBP using exact optimal control, DRL presents an alternative solution to address the problem. Once sufficiently trained in a simulated environment, DRL agents can provide real-time solutions by leveraging the forward pass of the neural network policy. Previous research has primarily focused on the ALBP in centralized settings, often tailored to specific types of assembly lines. Given the typically lengthy training times required for DRL, a generalized learning framework is necessary one that can adapt to diverse assembly line configurations without the need for retraining for each variation. Furthermore, techniques aimed at reducing training time would greatly enhance the efficiency and scalability of the training process. The main contributions of this work can be summarized as follows:
\begin{itemize}
	\item Unlike \cite{geurtsen2023deep}, this paper proposes a mathematical formulation of the GALBP within a DRL context, where a centralized environment interacts with multiple agents, rather than focusing solely on a serial assembly line.
	\item The decentralization challenge is addressed by transitioning from a single-agent to a multi-agent formulation, incorporating a sequential framework to enforce constraint satisfaction.
	\item To reduce training time, an action mask is employed to filter infeasible actions. This action mask is defined for both single-agent and multi-agent frameworks through a sequential feasibility check.
	\item A comparative analysis is conducted between DRL-trained agents and the OC approach, which serves as the baseline for evaluation.
\end{itemize}

\paragraph*{Structure} The remainder of the paper is organized into five sections. Section \ref{sec:Problem Formulation} presents the problem formulation. A brief description of the proposed masking technique is discussed in Section \ref{sec:Proposed Masking Technique}. Section \ref{sec:Multi Agent Decentralization Strategy} demonstrates the proposed decentralization strategy. Section \ref{sec:Results} illustrates the effectiveness of the proposed multi-agent scheme through numerical simulations. Finally, Section \ref{sec:Conclusion} concludes the work.

\section{Problem Formulation} \label{sec:Problem Formulation}
The mathematical formulation of the factory model requires defining state variables, the action space, a set of constraints, and the design of an optimization objective. Our focus will be on the problem formulation within the MDP framework. Specifically, a finite MDP is defined by the tuple $(S, A, p, \rho)$, where $S$ represents the set of states, $A$ denotes the set of actions, and $p(s,a,s')$ indicates the state-transition probability, which describes the likelihood that taking action $\mathbf{a[k]}$ from state $\mathbf{s[k]}$ leads to state $\mathbf{s[k+1]}$ (or $\mathbf{s'}$). The symbol $\rho$ represents the immediate reward received after the state transition resulting from taking action $\mathbf{a}$.

The RL scheduling problem must be formulated as a constrained MDP. Given the current state of the factory at time $k$, denoted $\mathbf{s[k]}$, not all actions are feasible, only those that satisfy a specific set of constraints. Therefore, a subset $\mathbf{A_{s}[k]} \subseteq A$ of the action space is admissible to ensure safe behavior. This necessitates the definition of a masking function $\mathcal{M}(s)$, which computes the admissible actions at a given state. The use of the masking function guarantees constraint satisfaction. In both proposed algorithms, our goal is to learn (via a neural network) the policy $\pi_{\theta_{k}}(s)$ that maximizes the expected cumulative reward, where $\theta_{k}$ represents the weights of the neural network at time $k$. In the case of on-policy PPO, we also need to learn a state-value function, which is approximated by a neural network with parameters $\phi_{k}$.

\subsection{Parameters and State Variables}
We denote the sets of positive integers and binary numbers as $\mathbb{Z}{+}$ and $\mathbb{Z}{2}$, respectively, with $|.|$ representing the cardinality of a set. The key parameters are as follows: the time horizon is $|K| \in \mathbb{Z}{+}$, the number of workstations is $|I| \in \mathbb{Z}{+}$, the number of tasks is $|J| \in \mathbb{Z}{+}$, and the number of different types of resources involved in the problem is $|R| \in \mathbb{Z}{+}$. The parameters related to the constraints are $\mathbf{O} \in \mathbb{Z}{+}^{|I|}$ and $\mathbf{U} \in \mathbb{Z}{+}^{|I| \times |R|}$, which represent the maximum capacities of the workstations in terms of the number of simultaneous tasks that can be processed and the amount of resources that can be stored, respectively. With respect to task constraints, $\mathbf{P} \in \mathbb{Z}_{2}^{|J| \times |J|}$ is a Boolean matrix representing time precedence constraints, $\mathbf{F} \in \mathbb{Z}{+}^{|J|}$ specifies the deadlines for each task, $\mathbf{D} \in \mathbb{Z}{+}^{|I| \times |J|}$ represents the task durations (which depend on the workstations), and $\mathbf{C} \in \mathbb{Z}{+}^{|J| \times |R|}$ indicates the amount of each specific resource needed for task completion. Finally, $\mathbf{G} \in \mathbb{Z}_{+}^{|R|}$ represents the overall availability of each resource type in the factory. The primary state variables of interest, which, when concatenated and flattened, represent the state of the factory, include the number of tasks currently being processed at the workstations (referred to as "occupancies") and denoted by $\mathbf{o[k]} \in \mathbb{Z}+^{|I|}$, the remaining durations for task completion, denoted $\mathbf{d[k]} \in \mathbb{Z}+^{|I| \times |J|}$, and the resources currently stored for task execution, denoted by $\mathbf{r[k]} \in \mathbb{Z}_+^{|I| \times |J| \times |R|}$. Other auxiliary variables are utilized to model the transitions of tasks between starting, execution, and completion states, as well as the availability of resources in both the factory and workstation storage. While additional features can be incorporated into the state vector when necessary, such additions would increase the vector’s size and require larger neural networks for the agents, thereby significantly increasing the training computation time. Therefore, careful consideration is required when adding new features to ensure they are essential for capturing the dynamics of the factory. The resulting state vector, formed by concatenation, has the following dimensionality: $\delta_s = \dim(\mathbf{s[k]}) = |I| \cdot (1 + |J| \cdot (1 + |R|))$.
\subsection{Factory Dynamics}
At the beginning of the simulation, unless the system is initialized in an intermediate state with pre-assigned tasks, the factory state is represented as a vector with $\delta_s$ zero entries. At time $k-1$, when the agent chooses to assign a task $j$ to workstation $i$, the corresponding $i$-th entry of $\mathbf{o[k]}$ is incremented by one, and this increment continues until the task is completed:
\begin{equation}
	o[k+1](i) =
	\begin{cases}
		o[k](i)+1 & \text{if } a[k](i,j) = 1 \\
		o[k](i) & \text{otherwise}
	\end{cases}
\end{equation}
The $(i,j)$-th entry of $\mathbf{d[k]}$ is then assigned to the corresponding initial duration value taken from $D$:
\begin{equation}
	d[k+1](i,j) =
	\begin{cases}
		D(i,j) & \text{if } a[k](i,j) = 1 \\
		d[k](i,j)-1 & \text{if } e[k](i,j) = 1 \\
		d[k](i,j) & \text{otherwise}
	\end{cases}
\end{equation}
After the assignment, at each time step, the duration is decreased by one, transitioning tasks from the execution state $\mathbf{e[k]} \in \mathbb{Z}_{2}^{|I| \times |J|}$ to the finishing state $\mathbf{f[k]}$. To minimize the search space, the resource action is kept binary, while the resources allocated for task execution, represented by the variable $\mathbf{r[k]}$, are integer-valued. This is enforced through the following dynamic constraint:
\begin{equation}
	r[k+1](i,j,r) =
	\begin{cases}
		r[k](i,j,r) + C(j,r) & \text{if } y[k](i,j,r) = 1 \\
		r[k](i,j,r) - C(j,r) & \text{if } d[k](i,j) = 0 \\
		r[k](i,j,r) & \text{otherwise}
	\end{cases}
\end{equation}
The resource assignment also corresponds to a reduction in factory inventories (i.e. $G[k+1](r)=G[k](r)-C(j,r)$). The resources assigned are considered task-specific since this improves traceability and allows modeling resources that do not vanish with consumption, such as tools.

%
%
\subsection{Actions Space}
\begin{figure*}[!htb]
	\centering
	\includegraphics[width=1\textwidth]{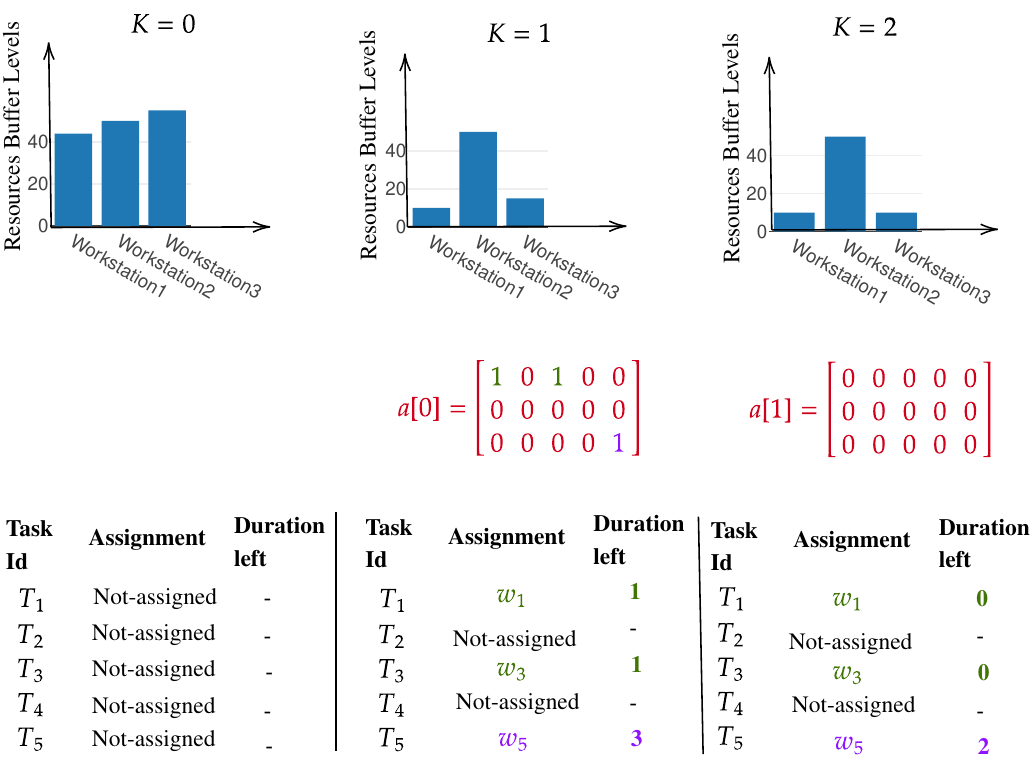}
	\caption{The visualization of the factory dynamics for a scenario involving 3 workstations and 5 tasks.}
	\label{fig: factory_dynamics}
\end{figure*}

\begin{figure*}[!htb]
	\centering
	\includegraphics[scale=0.75]{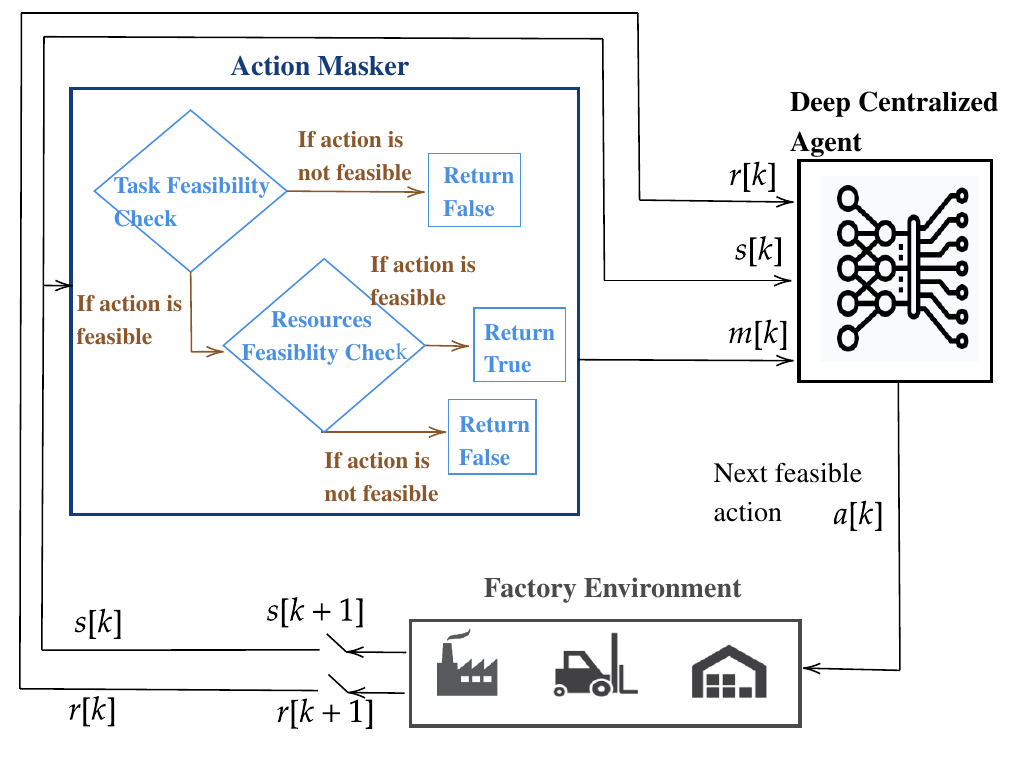}
	\caption{The central agent architecture which introduces a modified version of the DQN/PPO agents making use of the action mask concept.} \label{fig:scheme}
\end{figure*}

To solve the scheduling problem, dynamic assignment of tasks and resources to the workstations is required. These assignments define the actions within the problem. Let $\mathbf{a[k]} \in \mathbb{Z}_{2}^{|I| \times |J|}$ represent a Boolean matrix, where the $(i,j)$ entry equals one if task $j$ is assigned to workstation $i$ at time $k$. Similarly, let $\mathbf{y[k]} \in \mathbb{Z}_{2}^{|I| \times |J| \times |R|}$ denote a Boolean tensor, where the entry corresponding to resource $r$ and task $j$ at workstation $i$ at time $k$ is one if resource $r$ is assigned to task $j$ at workstation $i$. It is evident that the total number of possible actions, denoted as $\delta_A = |A|$, grows exponentially with the problem size.
\begin{equation}
	\delta_A = 2^{|I| \times |J|}
\end{equation}
The constraints inherent in the problem limit the number of feasible actions. Specifically, a task can be assigned to only one workstation at a time, and similarly, each resource can be allocated to a single task at a workstation. This imposes a summation constraint of one along the workstation dimension of the tensor. The number of actions that satisfy this unique assignment constraint is:
\begin{equation}
	\delta_A = (|I| + 1)^{|J|}
\end{equation}
In fact, each task can either be assigned to a workstation or remain unassigned. The number of feasible actions in the centralized framework is further constrained by the workstations' limited capacity to handle multiple simultaneous tasks, with the total number of tasks at any given workstation not exceeding its maximal occupancy. For the sake of simplicity, the maximal occupancy of workstation. 
It can be shown that the number of actions that satisfy both the unique assignment constraint and the workstation occupancy constraint is given by:

In fact, each task can be either assigned to one workstation or not assigned at all. The number of feasible actions in the centralized framework further reduces after taking into account the fact that 
the workstations have a limited capacity to handle multiple simultaneous tasks, whose total number can at most be equal to the specific workstation's maximal occupancy. 
It can be proved that the number of actions satisfying both the unique assignment constraint and the workstation occupancy constraint is given by:
\begin{equation} \label{eq:dim3}
	\begin{aligned} 
		\delta_A &= \sum_{i_1=0}^{O_1} \binom{|J|}{i_1} \sum_{i_2=0}^{\min\{O_2, |J|-i_1\}} \binom{|J|-i_1}{i_2} \\
		&\quad \sum_{i_3=0}^{\min\{O_3, |J|-i_1-i_2\}} \binom{|J|-i_1-i_2}{i_3} \\
		&\quad \sum_{i_{|I|}=0}^{\min\{O_{|I|}, |J|-i_1-\dots- i_{|I|-1}\}} \binom{|J|-i_1-\dots- i_{|I|-1}}{i_{|I|}}
	\end{aligned}
\end{equation}
where $O_i$ represents the i-th value of simultaneous occupancies limiting vector $O$.
This initial dimension is further reduced in the next step by introducing a decentralized framework. Furthermore, we simplify the problem by assuming that tasks and resource assignments are interlinked (i.e., if the 
$(i,j)$ component of $\mathbf{a[k]}$ equals one, then the $(i,j,r)$ components of the resource action $y[k]$ will also equal one, $\forall r$. Fig. \ref{fig: factory_dynamics} depicts  the factory dynamics for a simple case of 3 workstations and 5 tasks.  At time zero, the task allocation control matrix $a[0]$ assigns tasks 1 and 3 to workstation 1, while workstation 2 is not assigned any tasks, as indicated by the second row being zero in $a[0]$. Workstation 3 is assigned to task 5. In each subsequent time step, the remaining duration $d[1]$ decreases by one. Additionally, the resource inventory for workstation 1 and workstation 3 decreases accordingly. The dynamics of resource allocation follow a similar pattern, in alignment with the resource allocation matrix for each type of resource. At this specific time instant, the resource allocation matrix $y[k]$ is expected to assign resources to workstation 1 and workstation 3 in subsequent time steps, assuming the agent is effectively learning to manage factory resources.
The control scheme for the assembly line is shown in Fig. \ref{fig:scheme}.
%
%
%
\subsection{Constraints and Actions Masks}
The primary constraints in the problem include task deadlines, limited resource and storage availability, and precedence relationships among tasks. Additionally, logical constraints such as the inability to reassign completed tasks and the requirement for correct transitions between task states must also be enforced. Given the concatenated factory state $\mathbf{s[k]}$ and the previously outlined constraints, an action masking function maps the current state to a binary vector $\mathbf{m[k]}$ of the same dimension as the action space. In this vector, 0 represents infeasible actions, and 1 indicates feasible actions. The action masking function is formally defined as follows:
\begin{equation}
	\mathcal{\mathbf{M}}(s): \mathbb{Z}_{+}^{\delta_s} \longrightarrow \mathbb{Z}_{2}^{\delta_A}
\end{equation}
where the set of constraints used to do the mapping is defined in detail in Appendix A.
%
%

\subsection{Reward Function}
Formally defining an RL problem necessitates characterizing the reward function. In this work, we propose the following reward function: 
\begin{equation}
	\rho (a,s)[k] =  \beta \cdot \frac{h[k]}{1 + k^\alpha}.
	\label{eq:reward}
\end{equation}
where $\alpha$ and $\beta$ are positive scalar design parameters, and $\mathbf{h[k]}$ is a Boolean flag indicating whether all tasks have finished executing. The motivation behind the proposed reward function in \eqref{eq:reward} is to assign a higher reward for completing all tasks within fewer time steps, denoted by $k$. The absence of a reward in many episodes is not problematic, particularly when a fixed time horizon is considered, as the objective in reinforcement learning is to maximize the expected cumulative reward.
%
%
\section{Proposed Masking Technique}\label{sec:Proposed Masking Technique}
\begin{figure*}[!htb]
	\centering
	\includegraphics[scale=0.55]{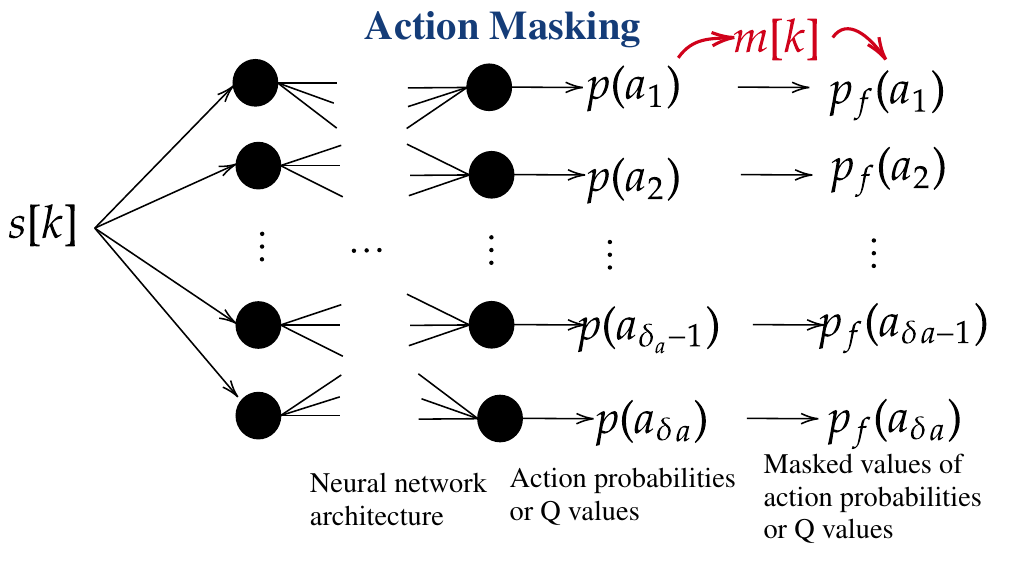}
	\caption{Action masking at the network output.} \label{fig: Action Masking1}
\end{figure*}

Action masking involves blocking any infeasible action suggested by the neural network agent and replacing it with the null action. As will be shown in the simulation section, action masking can significantly speed up the learning process. In this work, we have developed an action masking technique for both the off-policy DQN and the on-policy PPO algorithms, with only minor differences in their implementations. The process of action masking at the network's output is shown in Fig. \ref{fig: Action Masking1}. The details are explained in the following subsections.
%
%
\subsection{DQN Action Masking}
This paper presents a modified version of the DQN from \cite{mnih2013playing}, specifically adapted to incorporate action masking. In this approach, the policy network, which receives the centralized state as input, outputs $\delta_A$ Q-values. Prior to selecting the action with the highest Q-value, all Q-values corresponding to infeasible actions are set to negative infinity. As a result, the agent always selects feasible actions, even when a random action is chosen for exploration according to the well-known $\epsilon$-greedy policy. The masking technique also affects the loss computation, particularly when calculating the loss between predicted and target Q-values. In this process, the Q-values associated with infeasible actions are masked by setting them to $-\infty$.
%
%
\subsection{PPO Action Masking}
The PPO algorithm from \cite{schulman2017proximal} has been adapted in this work to integrate an action masking technique. The training process of a PPO agent involves the definition of two networks: an actor (or policy) network and a value network. The outputs of the actor network represent the probabilities for each possible action, given the current input state. In this approach, these probabilities are filtered by the action mask, which sets the probabilities corresponding to infeasible actions to negative infinity. The masked action probabilities are then re-normalized using the Softmax layer at the output, which is essential for multi-class classification tasks. As a result, infeasible actions are assigned null probabilities, while the probabilities of feasible actions are normalized so that they sum to one, as demonstrated in \cite{huang2020closer}. During the cost function optimization, the updated action probabilities, after applying the masking, are taken into account. One of the key advantages of using policy gradient methods with action masking is the stronger guarantee of convergence compared to action-value approximation. Specifically, the gradient update steps remain smooth, whereas action-value selection can lead to non-smooth gradient updates.
%
%
\subsection{Action Masking Effects}
The adoption of the mask drastically improves the convergence of the algorithm to the optimal policy. To show that, a PPO agent has been trained in the same environment and provided with a negative reward for taking infeasible actions.
\begin{figure}[!htb]
	\centering
	\includegraphics[scale=0.2]{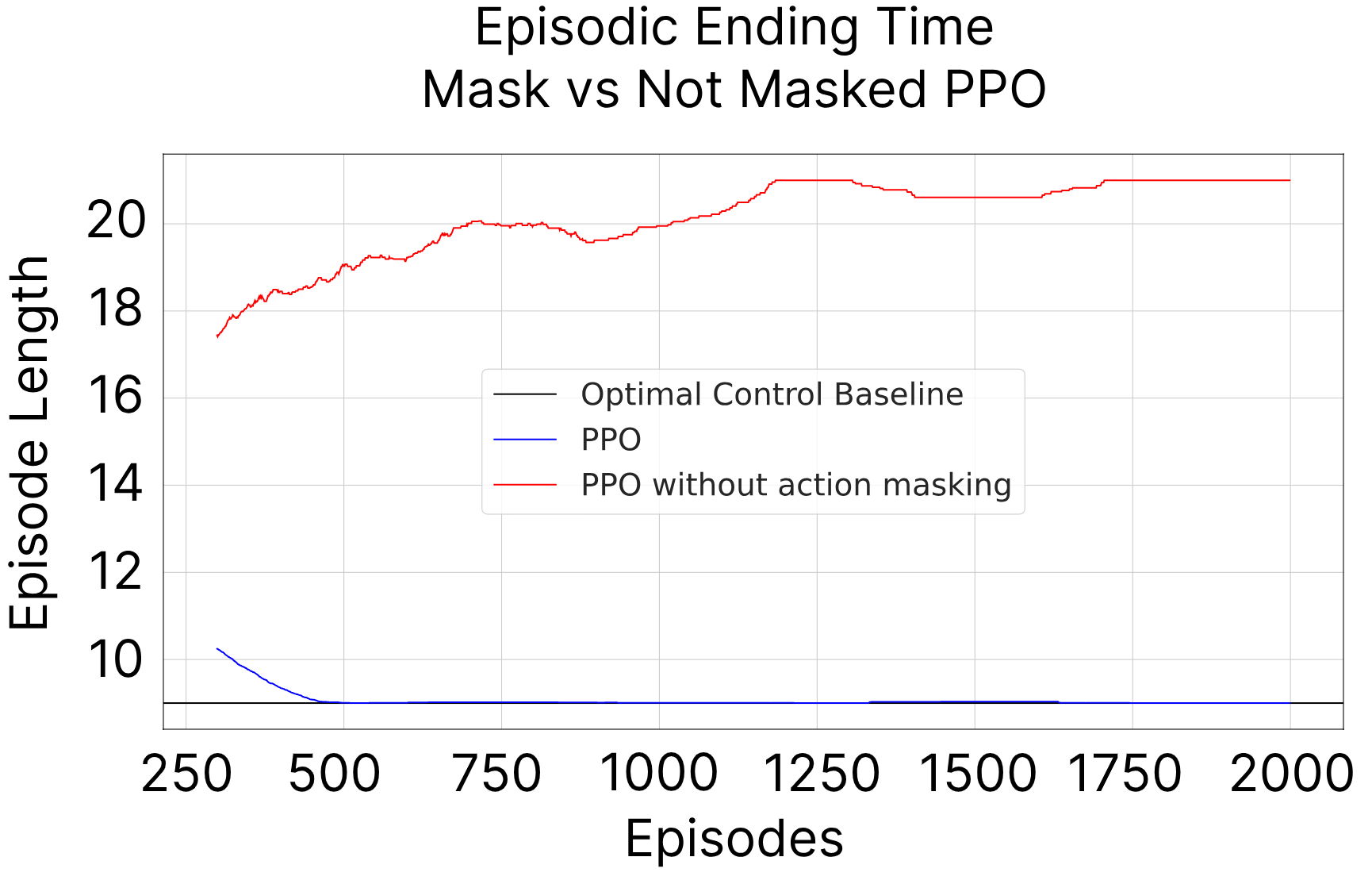}
	\caption{Action masking effect on the episodic length (finishing time of all tasks)  for centralized PPO algorithms compared to an equivalent optimal control problem in \cite{10174426}.}\label{fig: Mask Comp rh.}
\end{figure}
As an example, Fig.  \ref{fig: Mask Comp rh.} shows the comparison of the episodic length (finishing time of all tasks) between a masked-PPO and the classical one, highlighting the effectiveness of the strategy. In the latter case, there is no convergence to the optimal policy, whereas the masked version achieves an optimal solution within only 400 training episodes.
%
%
\section{Problem Decentralization Strategy and Proposed Series Constraints Enforcement Scheme} \label{sec:Multi Agent Decentralization Strategy}
The dependence of the observation and action space dimensions on factory parameters, and the resultant dimension explosion, must be considered to maintain scalability. Bearing this in mind, RL possesses a suitable structure for learner decentralization, thereby reducing computational complexity. In our scenario, the idea is to decentralize the assignment at the level of the workstations. We can consider each workstation as an agent aiming to solve a cooperative task. The constraints satisfaction plays an important role; in fact, logic must be implemented to guarantee that the actions selected by the agents are not in conflict with each other. Therefore, a more robust control logic is required to efficiently ensure that the concatenated action, resulting from the decentralized agents' choices, is feasible. This problem has been solved through a sequential feasibility check that makes use of the masks.
%
%
\subsection{Multi-agent Formulation}
The multi-agent problem requires decentralization of the observations space, the actions space, the actions masking, and the design of a new reward function. 
The assumption is made that the controlled system is fully observable, which is reasonable in the case of a factory. The centralized fully observable state will be therefore maintained. The action space has been decentralized by considering the specific occupancy constraint for the $i-th$ workstation. This imposes a summation constraint on each action, ensuring that workstations cannot assign themselves more tasks than their occupancy bounds allow. The following decentralized dimensions are obtained:
\begin{equation}
	\delta_{A_{i}} = \sum_{j=0}^{O_{i}} \binom{|J|}{j}, \ \ \ \    \forall i
\end{equation}  
This implies that the $|I|$ decentralized mappings will have varying co-domain dimensions:
\begin{equation}
	\mathcal{\mathbf{M}}_i(s): \mathbb{Z}_{+}^{\delta_s} \longrightarrow \mathbb{Z}_{2}^{\delta_{A_{i}}}, \ \ \ \    \forall i
\end{equation}
The centralized action for task assignment $\mathbf{a[k]}$ can be redefined by splitting it row-wise. Each row consists of task assignments to a specific workstation, enabling us to redefine each row as the decentralized action $\mathbf{a_i[k]}$. The reward function employed for this problem is consistent with the one used in the centralized case, and it is shared among all agents. Furthermore, all agents have access to the same observations.
%
%
\subsection{Sequential Feasibility Check}
\begin{figure}[!htb]
	\centering
	\includegraphics[scale=0.55]{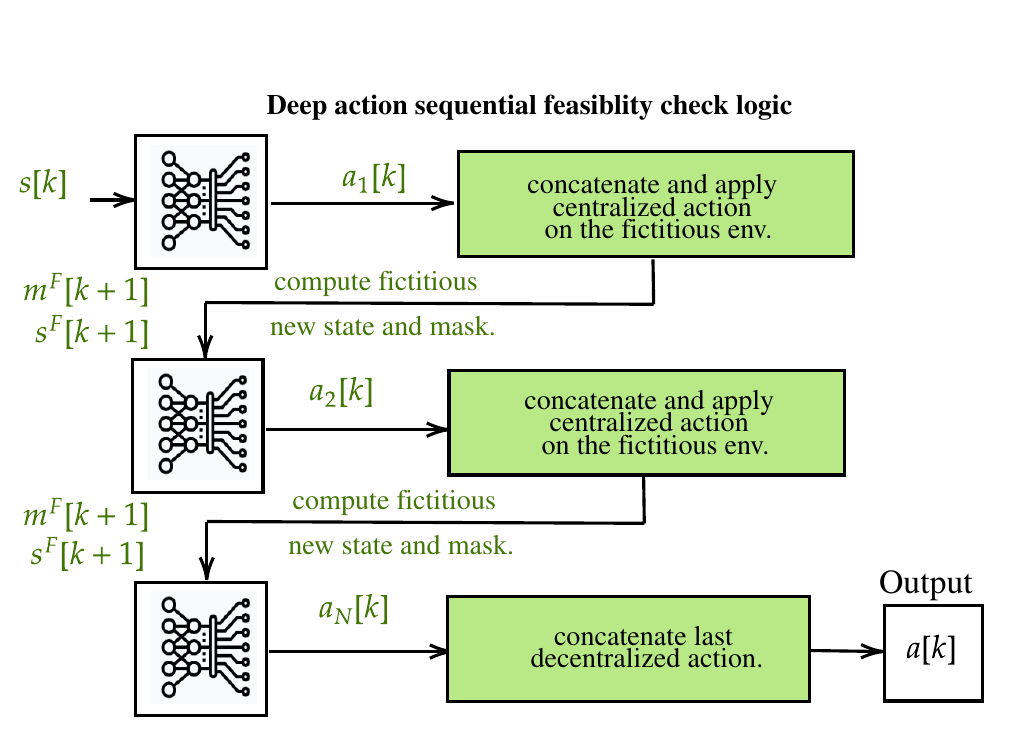}
	\caption{Sequential action feasibility check control logic.}\label{fig: SFC}
\end{figure}

Sequential actions taken independently by the agents might result in an infeasible aggregated action. For this reason, the same initial state is fed to all the agents, outputting their predicted actions. Then these actions are concatenated sequentially into a matrix with the same shape as the centralized action, and its feasibility is checked. If the action is not feasible, it is surely due to a conflict between the agents' choices, necessitating a feasibility check. This is done by iteratively selecting a random agent and applying its proposed action to a simulated copy of the environment. Then the resulting new simulated state is evaluated. The fictitious state $s^F$, is then applied as input to the next agent in the sequence, which will compute the next best action to be applied, based on a mask computed from the fictitious state, defined as $m^F$. This loop will proceed sequentially for all the $|I|$ agents, outputting a feasible concatenated action. The main idea of this loop is to provide the agents with a copy of the real state of the factory, where assets are allocated following agents' choices but without decreasing the duration of the tasks (e.g., without executing them). Based on this fictitious state, a mask is computed. The mask can be considered a fictitious mask because it is computed dependently on the choices of the previous agents in the loop. (e.g., if the first agent in the loop assigns a task, the fictitious mask $m^F$ computed by the next agent will not see that task as an assignable one, even if it is). To prevent biases, the order is randomized during learning. The use of this control loop allows the agents to be "conscious" of the behavior of the others, guaranteeing an efficient action selection in a decentralized way. Agents in this loop are randomly selected during training to avoid biased results. The sequential feasibility check control scheme is shown in Fig.  \ref{fig: SFC}.
The sequential feasibility check procedure is detailed in Algorithm \ref{alg:Sequential Feasibility Check}, in which the following functions are defined:
\begin{itemize}
	\item \textbf{InitFakeEnv}$\mathbf{(s^F[k], m^F[k])}$: this function is used to initialize the fictitious environment.
	\item \textbf{FakeTransition}$\mathbf{(a[k], s^F[k], \mathcal{E}^F)}$: it is used to update the current simulated state and masks as a result of the sequential actions proposed by each agent. These transitions of the simulated states are used to check the feasibility of the proposed concatenated action, computed sequentially (they do not cause a real state transition).
\end{itemize}
The dependencies from time $k$ will be omitted in the following algorithm for the sake of clarity (e.g., $s[k]$ and $m[k]$ will be treated as $s$ and $m$).
\begin{algorithm}[!htb]
	\caption{Sequential Feasibility Check (Randomized)}\label{alg:Sequential Feasibility Check}
	\begin{algorithmic}[1]
		\Require Environment $\mathcal{E}$, Agents $\mathcal{A}_i$
		\Require Current State $s[k]$
		\Require Current Mask $m = [m_1, \dots, m_{\delta_{A_i}} ]$
		\Require Initial Feasible Action $a = [\dots, \ 0,  \ \dots]$
		\State $s^F = s$, $m^F = m$
		\State $\mathcal{E}^F=\text{InitFakeEnv}(s^F, m^F)$
		\For{$i = 1$, $\dots$,$|I|$}
		\State Choose a random agent $\mathcal{A}_i$
		\State $a_i = \text{GetAgentAction}(s^F, m^F_i,\mathcal{A}_i)$
		\State $a = [\dots,   a_i,  \dots]$
		\State $s^F, m^F = \text{FakeTransition}(a, s^F, \mathcal{E}^F)$
		\EndFor
		\State \Return $\mathbf{a[k]}$
	\end{algorithmic}
\end{algorithm}
\subsection{Multi Agent Training Loop}
The pseudocode for the MARL training loop is presented in Algorithm \ref{alg:MARL Training}. The main functions defined and used are:
\begin{itemize}
	\item \textbf{Reset}($\mathcal{E}$): it is the reset function of the environment. It returns the episodic initial state.
	\item \textbf{GetAgentMask$\mathbf{(s[k], \mathcal{A}_{i}, \mathcal{E})}$}: is called on the environment to get a mask depending on the current state, for each agent.    
	\item \textbf{GetAgentAction}$\mathbf{(s[k], m_i[k], \mathcal{A}_{i})}$: this function is called on each agent to output the predicted action using the current policy.
	\item \textbf{GetStateValue}$\mathbf{(s[k], \mathcal{A}_{i})}$: is called on each agent to get the value function for the given state, for each agent.
	\item \textbf{SFC}$\mathbf{(s[k], a[k], m[k])}$: is the main control logic used to create a feasible concatenated action once we detect an infeasibility source, applying Algorithm \ref{alg:Sequential Feasibility Check}.
	\item \textbf{Transition}$\mathbf{(a[k], \mathcal{E})}$ is the state transition function of the environment.
	\item \textbf{ComputeProbsNewAction}$\mathbf{(s[k], a_i[k], m_i[k], \mathcal{A}_{i})}$: it is used if we change an infeasible action proposed by the agent to a safe one. It computes the probabilities of the new safe action for the memory update.
	\item \textbf{MemoUpdate} $\mathbf{(s[k], a_i[k], p_i[k], v_i[k], r[k+1],}$ \\
	$\mathbf{h[k+1], m_i[k])}$:
	This function is used to update the memory buffer of each agent and collect safe state trajectories using the mask embedding. Here, $p_i[k]$ refers to a specific element of $p_{f}$ for the $i$-th agent, depending on the selected action $a_i[k]$.
	\item \textbf{MaskedPPOLearningStep}$\mathbf{(\mathcal{A}_i)}$: indicates the learning steps of Masked PPO-Clip.   
\end{itemize} 
\begin{algorithm}[!htb]
	\caption{Multi-agent PPO Training Loop} \label{alg:MARL Training}
	\begin{algorithmic}[1]
		\Require Environment $\mathcal{E}$
		\Require PPO Agents $\mathcal{A}_{i}$, for $ i \in 1,..., |I|$
		\For {$n_{\text{episodes}} = 1, \dots , N_{\text{episodes}}$}
		\State $s$ = Reset($\mathcal{E}$)
		\While {not done}
		\For$i = 1, \dots, |I|$
		\State $m_i = \text{GetAgentMask}(s, \mathcal{A}_i, \mathcal{E})$
		\State $a_i[k]$ = $\text{GetAgentAction}(s, m_i, \mathcal{A}_i)$
		\State $v_i = \text{GetStateValue}(s, \mathcal{A}_i)$
		\EndFor
		\State $a = [ a_1, \dots, a_{|I|} ] $
		\State $m = [ m_1, \dots, m_{|I|} ] $
		\If    $a$ is not feasible
		\State $a = \text{SFC}(s, a, m)$
		\EndIf
		\State $s'$, $r$, $h  = \text{Transition}(a, \mathcal{E})$
		\For{$i = 1, \dots, |I|$}
		\State $p_{i} = \text{ComputeProbsNewAction}(s, a_i, m_i, \mathcal{A}_i)$
		\State $\text{MemoUpdate}(s, a_i,p_i, v_i, r, h, m_{i,t})$
		\EndFor
		\If {$f_{\text{learn}} = \text{True}$}
		\State $\text{MaskedPPOLearningStep}(\mathcal{A}_i)$,$\forall i$
		\EndIf
		\State $s \gets s'$
		\EndWhile
		\EndFor
	\end{algorithmic}
\end{algorithm}
Algorithm \ref{alg:MARL Training} shows the training loop for the multi-agent system. The time dependencies have been omitted for the sake of clarity. It starts by initializing the environment of the factory and the number of agents, whereas for each episode it starts by resetting the environment, and while not all the tasks are finished it proceeds with the main while loop. In the main while Loop in step 3 for each agent, the algorithm computes the current action mask and builds the concatenated action based on the policy of each agent. Then in step 11, the feasibility of the concatenated action is checked, and if it is not feasible, a feasible action is created in step 12. In step 14, the feasible concatenated action is applied to the environment. $s'$ is a simplified notation for $s[k+1]$. Subsequently, the memory is updated. Learning, based on the masked version of PPO, continues until all tasks are completed. For DQN, the sole modification to Algorithm \ref{alg:MARL Training} starts from step 15, where learning is based on the masked version of DQN rather than PPO. The proposed framework work falls under centralized training and decentralized execution within a multi-agent framework. Centralized training and decentralized execution have been shown to effectively address challenges in wireless power allocation \cite{kopic2024collaborative} and power control in heterogeneous networks \cite{zhang2020deep}. The primary reason for adopting this strategy in the proposed work is the implementation of a sequential feasibility check, which determines whether an action at a given workstation is feasible. This feasibility check must be centralized to ensure, for instance, that a task has not already been assigned by another workstation.
%
\section{Results} \label{sec:Results}
This section presents the main results and proposes some empirical tuning guidelines. All simulations were conducted on an Intel(R) Core(TM) i7-10750H CPU @ 2.60GHz, 2592 MHz, 6 cores, 16 GB RAM, and a 64-bit processor running Windows 11. The key parameters of the factory and the agents are summarized in Tables \ref{tab1}, \ref{tab2}, and \ref{tab3}. These parameters were fine-tuned through multiple experiments to identify the optimal configurations. Special attention should be given to tuning the target synchronization frequency in DQN and the learning frequency in PPO. Although a grid search over the parameters could be conducted for further optimization, this was deferred to future work due to computational constraints. Empirically, good performance was observed when the network's input layer size was set to 3 to 5 times the input size $\delta_s$. In general, PPO exhibited less sensitivity to hyperparameters compared to DQN. For both PPO and DQN, the update frequency is the most crucial parameter, as it influences both the stability and efficiency of the learning process. The remaining parameters in Table \ref{tab1} are relatively straightforward to tune. A smaller value for the learning frequency and target synchronization frequency than those specified in Table \ref{tab1} may lead to more frequent updates between the Q-network and the target network, which can cause instability in the learning process. Conversely, larger values can enhance stability but may result in slower learning improvements.
\begin{table*}[hbt!]
	\begin{center}
		\caption{Agent Parameters.}
		\begin{adjustbox}{width=1\textwidth}
			\begin{tabular}{|c |c| c| c| c|} 
				\hline
				\label{tab1}
				Hyperparameters & DQN Centralized & DQN multi-agent & PPO Centralized & PPO multi-agent \\ [0.5ex] 
				\hline \hline
				Learning Rate (BOTH) & 1e-5 & 1e-5 & 3e-4 & 3e-4 \\ 
				\hline
				Batch Size (BOTH) & 64 & 64 & 5 & 5\\
				\hline
				Discount (BOTH) & 0.995 & 0.995 & 0.99 & 0.99 \\
				\hline
				Memory (BOTH) & 100000 & 100000 & 100000 & 100000  \\
				\hline
				Target Sync Frequency (DQN) & 10 & 10 & - & -  \\
				\hline
				Soft Update Weight (DQN) & 0.8 & 0.8 & - & - \\
				\hline
				Gradient Clipping Value (DQN) & 1 & 1 & - & - \\
				\hline
				Learning Warm start (DQN) & 200 & 200 & - & - \\
				\hline
				Likelihood Ratio Clipping (PPO) & - & - & 0.2 & 0.2 \\
				\hline
				Learning Frequency (PPO) & - & - & 20 & 20 \\
				\hline
				Learning Epochs (PPO) & - & - & 4 & 4 \\
				\hline
				GAE lambda value (PPO) & - & - & 0.95 & 0.95 \\
				\hline
			\end{tabular}
		\end{adjustbox}
	\end{center}
	\vspace{-4mm}
\end{table*}

\begin{table*}[hbt!]
	\begin{center}
		\caption{Networks Parameters.}
		\begin{adjustbox}{width=1\textwidth}
			\begin{tabular}{|c| c| c| c| c|}
				\hline
				\label{tab2}
				Network Shape & DQN & DQN Multi-agent & PPO & PPO Multi-agent \\ [0.5ex]
				\hline\hline
				Input Layer Neurons  & $\delta_s$ & $\delta_s$ & $\delta_s$& $\delta_s$  \\
				\hline
				Input Layer Activation Function  & Tanh & Tanh & Tanh & Tanh \\
				\hline
				First Hidden Layer Neurons & 534 & 178 & 258 & 86 \\ 
				\hline
				First Hidden Layer Activation Function & Tanh & Tanh & Tanh & Tanh\\
				\hline
				Second Hidden Layer Neurons & 534 & 178 & 258 & 86 \\
				\hline
				Second Hidden Layer Activation Function & Tanh & Tanh & Tanh & Tanh  \\
				\hline
				Third Hidden Layer Neurons  & 534 & 178 & 258 & 86  \\
				\hline
				Third Hidden Layer Activation Function  & Tanh & Tanh & Tanh & Tanh \\
				\hline
				Output Layer Neurons (i-th agent)  & $\delta_{A}$& $\delta_{A_{i}}$ & $\delta_{A}$ & $\delta_{A_{i}}$  \\
				\hline
				Output Layer Activation Function  & - & - & Softmax & Softmax \\
				\hline
			\end{tabular}
		\end{adjustbox}
	\end{center}
	\vspace{-4mm}
\end{table*}
\begin{table}[hbt!]
	\begin{center}
		\caption{Factory  Parameters.}
		\label{tab3}
		\begin{tabular}{|c |c|} 
			\hline
			Parameter & Value \\ [0.5ex] 
			\hline\hline
			$|K|$& 20 \\ 
			\hline
			$|I|$ & 3 \\
			\hline
			$|J|$ & 5 \\
			\hline
			$|R|$ & 2 \\
			\hline
			$O$ & $\begin{bmatrix} 1 & 3 & 1
			\end{bmatrix}$ \\
			\hline
			$U$ & $\begin{bmatrix} 50 & 50 \\
				50 & 50 \\
				50 & 50                               
			\end{bmatrix}$ \\
			\hline
			$D$ & $\begin{bmatrix} 4 & 5 & 8 & 5 & 2 \\
				5 & 6 & 12 & 3 & 6 \\
				3 & 4 & 5 & 3 & 5 
			\end{bmatrix}$ \\
			\hline
			$F$ & $\begin{bmatrix} 20 & 20 & 20 & 20 & 20\end{bmatrix}$ \\
			\hline
			$C$ & $\begin{bmatrix} 13 & 12 \\
				6 & 7 \\
				4 & 3 \\
				2 & 4 \\
				3 & 5
			\end{bmatrix}$  \\
			\hline
			$P$ & $\begin{bmatrix} 0 & 1 & 0 & 0 & 1 \\
				-1 & 0 & 0 & 0 & 0 \\
				0 & 0 & 0 & 0 & 0 \\
				0 & 0 & 0 & 0 & 1 \\
				-1 & 0 & 0 & -1 & 0
			\end{bmatrix}$  \\
			\hline
			$G$ & $\begin{bmatrix} 1000 & 2000 \end{bmatrix}$   \\
			\hline
		\end{tabular}
	\end{center}
\end{table}
\subsection{Agents comparison}
\begin{figure}[!htb]
	\centering\includegraphics[scale=0.17]{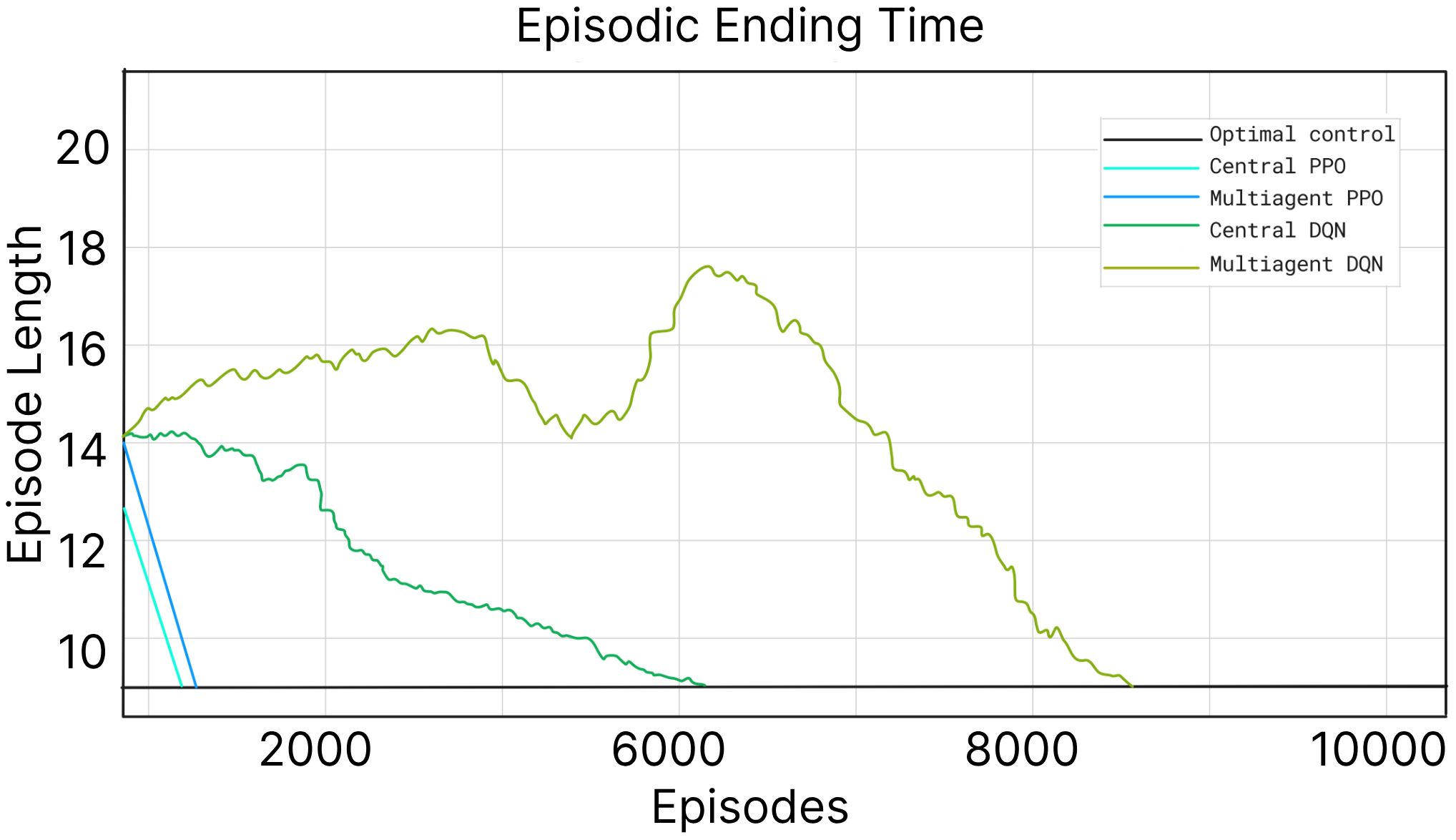}
	\caption{ Comparison of the episodic ending times (time necessary to complete all tasks) during training.} 
	\label{fig: Comp eh}  
\end{figure}

The training processes of the agents were compared, with the solution to the problem formulated within the optimal control (OC) framework serving as a benchmark for performance evaluation. The agents were tuned to converge to the same outcomes as the equivalent OC problem. The training time for the PPO agent was found to be the shortest overall, even with a larger network. Additionally, the decentralization of the problem justifies the use of a smaller network size, as it leads to smaller action spaces and, consequently, reduces training time by up to $50\%$. In Fig. \ref{fig: Comp eh}, the episodic completion times achieved during training for both the centralized and decentralized versions of DQN and PPO are depicted, with the solution obtained from OC shown in black. 
The y-axis represents the achieved $k_{\text{end}}$ in each episode, where $k_{\text{end}}$ denotes the time instance when all tasks are completed, or when the environment reaches the horizon value. The objective in DRL is to achieve the minimum possible ending time $k_{\text{opt}}$, which is computed by solving an optimization problem.
%

%
%
\subsection{Robustness Empirical Test to a Random Initialization of the Trained PPO Agent}

\begin{figure}[!htb]
	\centering
	\begin{subfigure}[b]{0.3\textwidth}
		\centering
		\begin{tikzpicture}
			\small
			\pie[radius=1.5, color={red!70, blue!70}, text=, rotate=215]%
			{5.47/Sub-Optimal Solutions, 95.53/Optimal Solutions}
			\pie[hide number, radius=1.7, color={red!70, blue!70}, rotate=215]%
			{5.47/\!5.47\%}
		\end{tikzpicture}
		\caption{$i=3, j=5$}
	\end{subfigure}
	\begin{subfigure}[b]{0.3\textwidth}
		\centering
		\begin{tikzpicture}
			\small
			\pie[radius=1.5, color={red!70, blue!70}, text=, rotate=215]%
			{7.89/Sub-Optimal Solutions, 92.11/Optimal Solutions}
			\pie[hide number, radius=1.7, color={red!70, blue!70}, rotate=215]%
			{7.89/\!7.89\%}
		\end{tikzpicture}
		\caption{$i=15, j=10$}
	\end{subfigure}
	\begin{subfigure}[b]{0.3\textwidth}
		\centering
		\begin{tikzpicture}
			\small
			\pie[radius=1.5, color={red!70, blue!70}, text=, rotate=215]%
			{8.13/Sub-Optimal Solutions, 91.87/Optimal Solutions}
			\pie[hide number, radius=1.7, color={red!70, blue!70}, rotate=215]%
			{8.13/\!8.13\%}
		\end{tikzpicture}
		\caption{$i=10, j=15$}
	\end{subfigure}
	
	\vspace{0.5cm}
	\begin{tikzpicture}
		\node[rectangle, draw, fill=red!70, minimum width=1cm, minimum height=0.5cm] at (0,0) {};
		\node[right] at (0.7,0) {\small Sub-Optimal Solutions};
		\node[rectangle, draw, fill=blue!70, minimum width=1cm, minimum height=0.5cm] at (0,-1) {};
		\node[right] at (0.7,-1) {\small Optimal Solutions};
	\end{tikzpicture}
	
	\caption{Robustness test results.}
	\label{fig: robustness}
\end{figure}
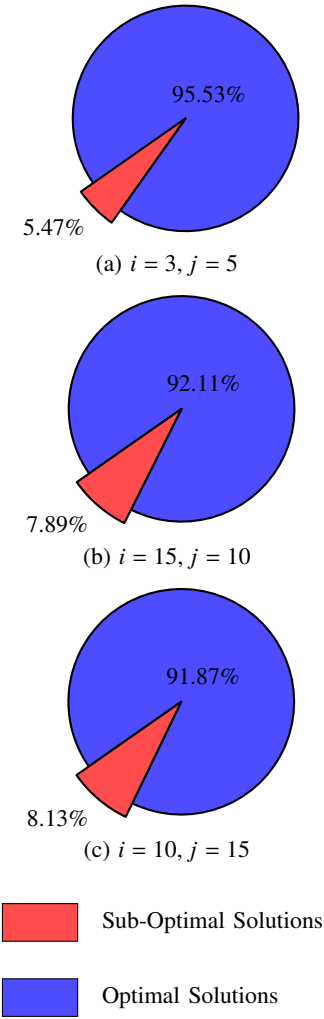

To evaluate the robustness of the proposed model, an empirical test was conducted. In this test, a pre-trained PPO agent was initialized with a set of random initial states. The minimum achievable ending time for each random state was computed and compared with the actual ending time achieved by the agent. The proportion of optimal solutions obtained without further learning is presented in Fig. \ref{fig: robustness}. The PPO agent successfully provided optimal solutions in over $90\%$ of the instances across different values of $i$ and $j$. These results highlight the potential of the proposed approach to compute on-the-fly solutions for the ALBP, even when initialized with random states. It is worth noting that the PPO agent is well-regarded for its robustness to random initialization, as demonstrated in previous works \cite{emami2024age} and \cite{ali2024deep}.

%
%
%

%
%
\subsection{Multi-Agent Results}

\begin{figure*}[!htb]
	\centering
	\begin{subfigure}[b]{0.425\textwidth}
		\centering
		\includegraphics[scale=0.14]{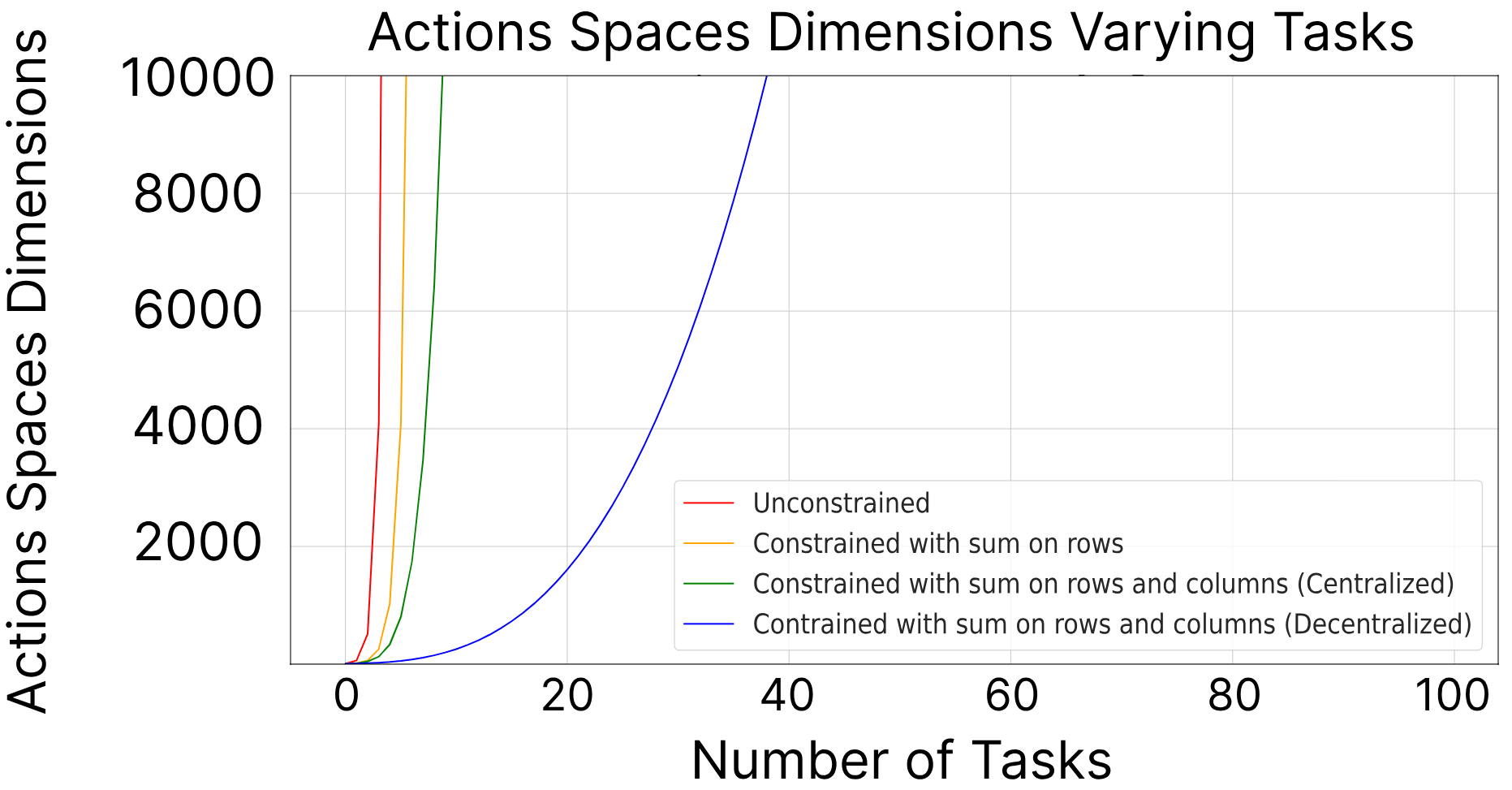}
		\caption{Actions space dimensions varying the number of task.}
		\label{fig: actions space dimension comparison}
	\end{subfigure}
	\hfill
	\begin{subfigure}[b]{0.425\textwidth}  
		\centering 
		\includegraphics[scale=0.14]{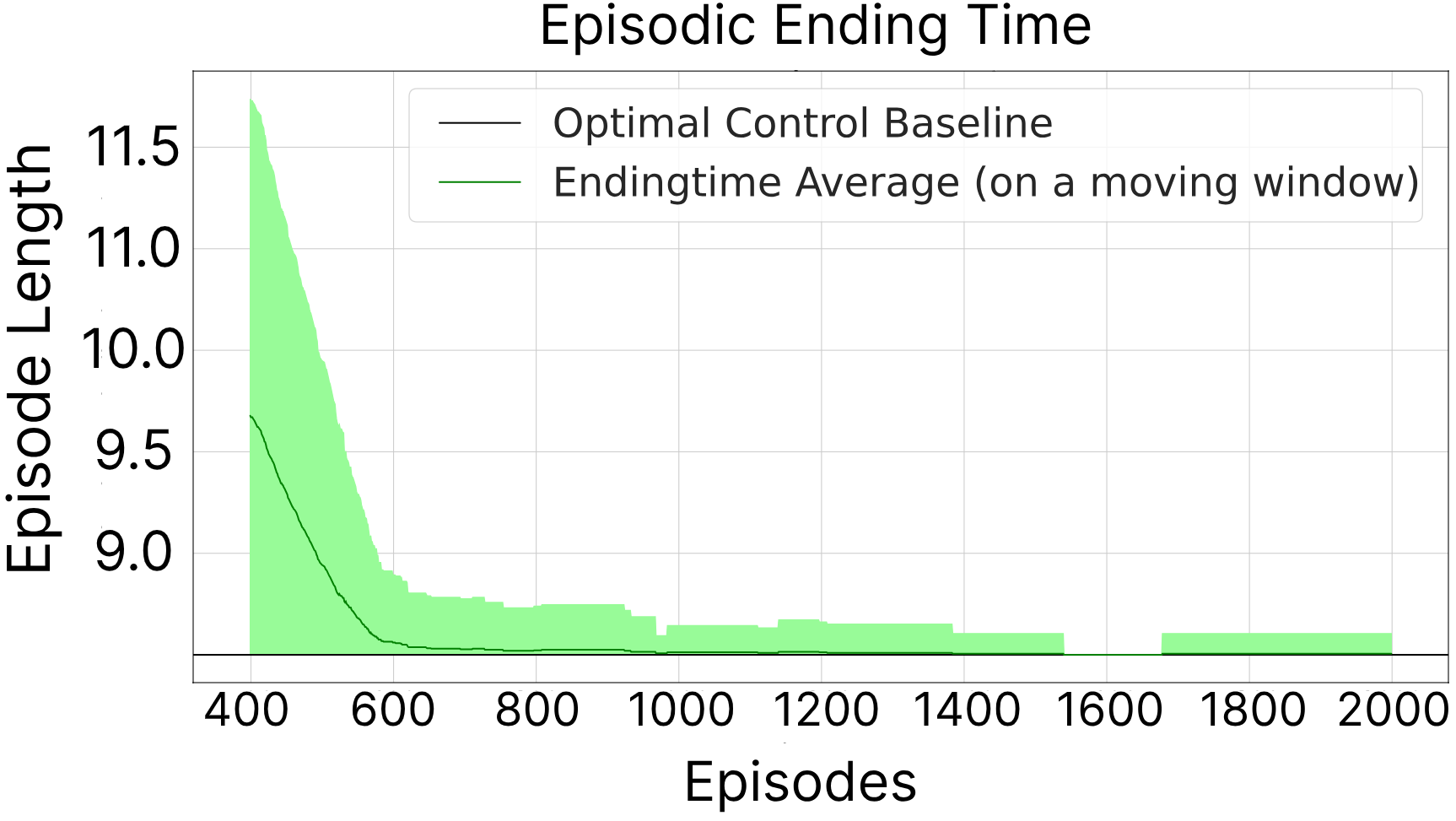}
		\caption{PPO Multi-agent episodic ending times during training.} \label{fig: PPOm eh1}
	\end{subfigure}
	\vskip\baselineskip
	\begin{subfigure}[b]{0.425\textwidth}  
		\centering 
		\includegraphics[scale=0.14]{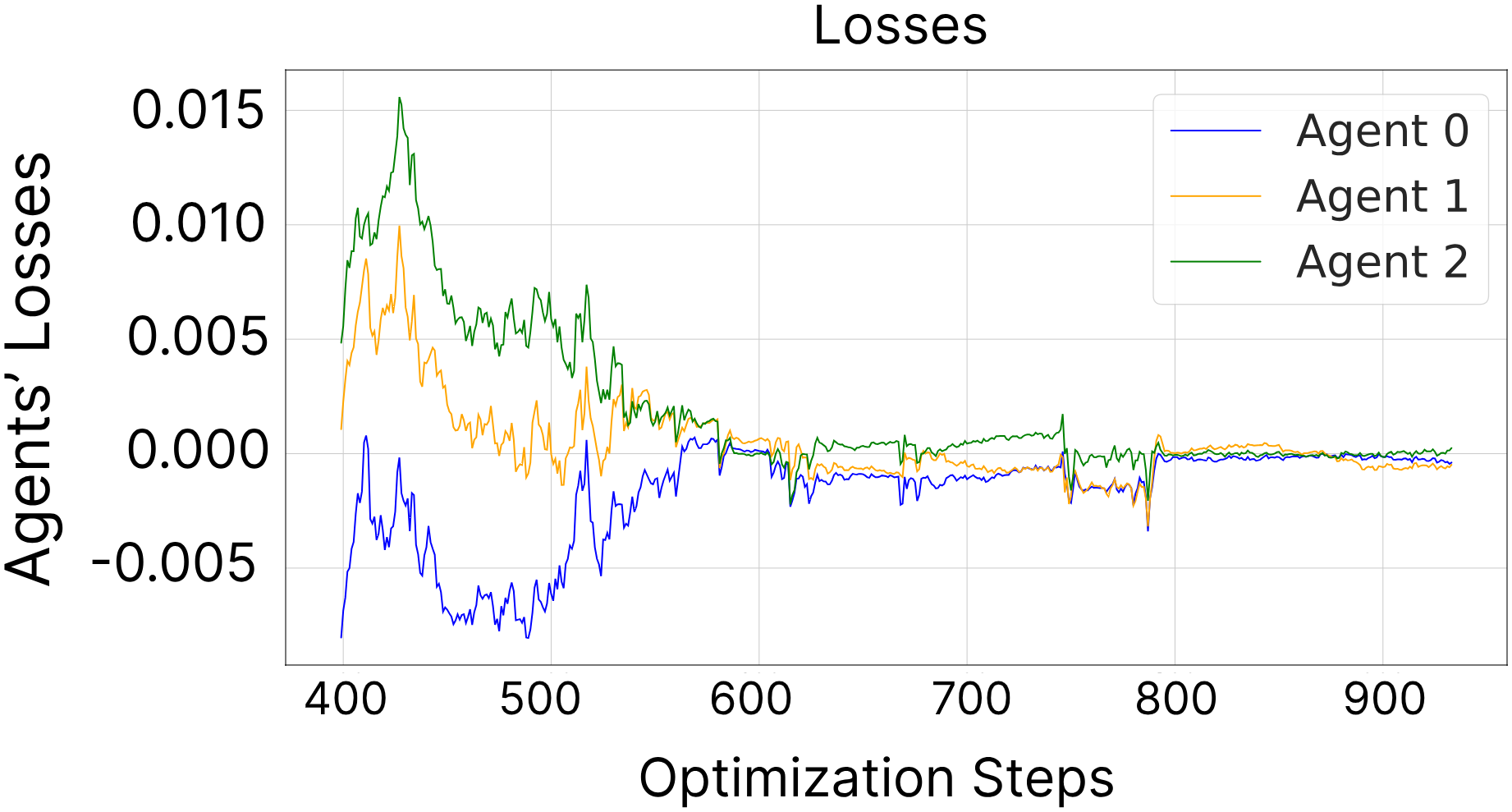}
		\caption{PPO Multi-agent losses during training.} \label{fig: PPOm eh}
	\end{subfigure}
	\hfill
	\begin{subfigure}[b]{0.425\textwidth} 
		\centering 
		\includegraphics[scale=0.14]{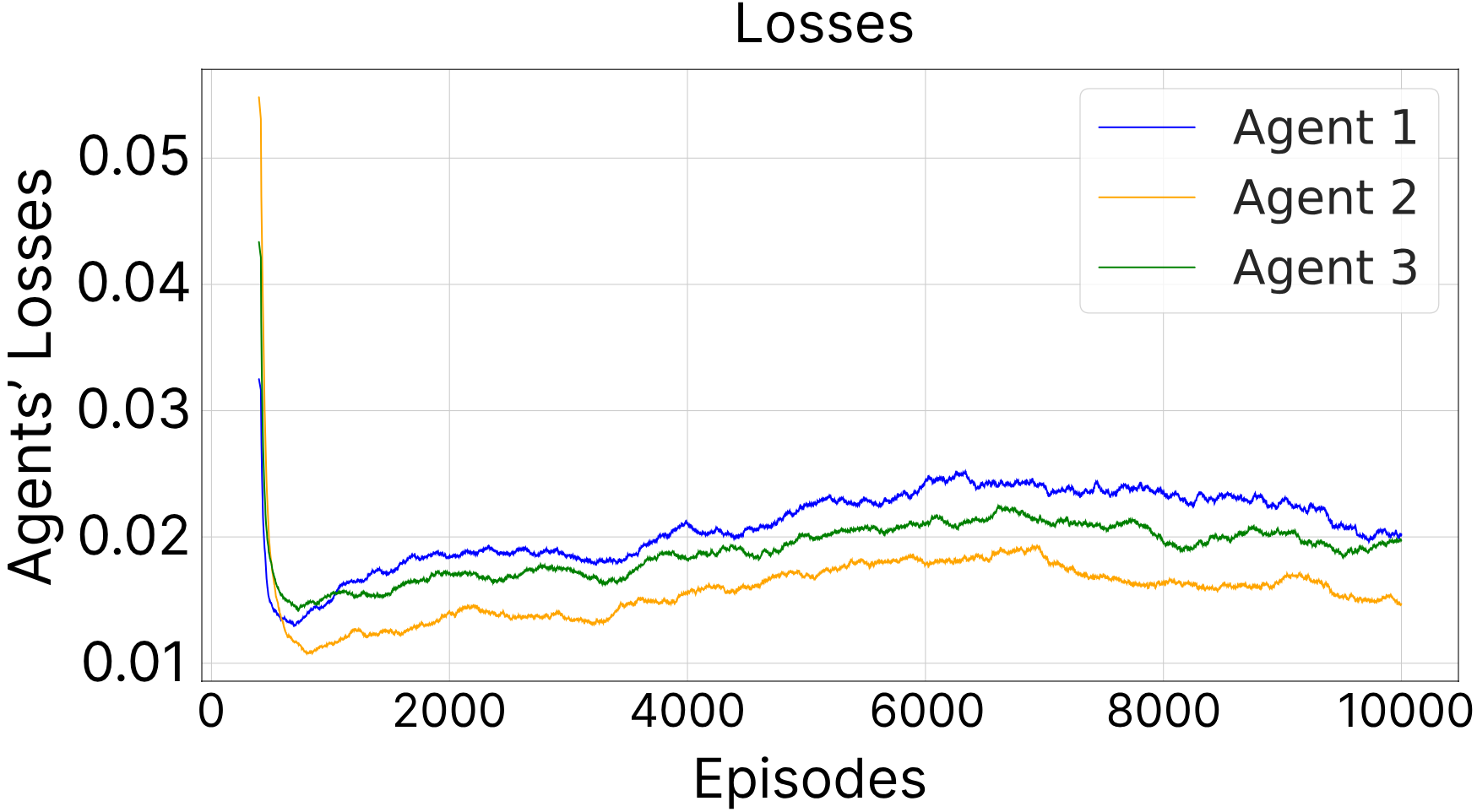}
		\caption{DQN Multi-agent losses during training.} \label{fig: DQNm eh}
	\end{subfigure}
	\caption{Part (a) shows how the action space size increases, varying the number of tasks. Part (b) illustrates the episodic ending time achieved by centralized PPO during training. Part (c) and Part (d) show the agents' losses during training for multi-agent PPO and DQN, respectively.}
	
	\label{fig:Result1}
\end{figure*}
Fig. \ref{fig: actions space dimension comparison} illustrates the growth in the dimension of the action space as the number of tasks increases, for the various scheduling algorithms discussed in this paper. The figure highlights the polynomial dependence achieved by the proposed multi-agent framework, in contrast to the initial exponential growth characteristic of the single-agent framework. This result opens avenues for future research, such as further decentralizing the task assignment process, which could further reduce the dimension of the action space. The episodic ending time achieved by the multi-agent PPO during training is presented in Fig. \ref{fig: PPOm eh1}. The shaded areas represent the standard deviation around the average ending times, computed using a moving window. The plot demonstrates that the multi-agent system converged to the optimal policy in very few iterations, despite being trained with smaller networks compared to the centralized network. As shown in the subsequent plots, the multi-agent system also achieved convergence in terms of loss during the training process. Table.  \ref{tab5} summarizes the main training results. It is evident that the proposed multi-agent PPO significantly reduces the training time compared to both the centralized and multi-agent DQN approaches, achieving a nearly $50\%$ reduction in training time relative to the centralized PPO.
\begin{table*}[!tbh]
	\centering{}\caption{\label{tab5}Training results comparison}
	\begin{tabular}{|>{\raggedright}p{2.6cm}|c|>{\centering}p{1.3cm}|>{\centering}p{2cm}|>{\centering}p{1.4cm}|}
		\hline 
		\multirow{2}{2.6cm}{\centering{}Case } & \multirow{2}{*}{Algorithm } & \multirow{2}{1.3cm}{\centering{}Training Time} & \multirow{2}{2cm}{\centering{}Convergence Episode} & \multirow{2}{1.4cm}{\centering{}Inference Time}\tabularnewline
		&  &  &  & \tabularnewline
		\hline 
		\hline 
		\multirow{5}{2.6cm}{3 Workstations, 5 Tasks} & DQN Centralized  & 01:05:48  & 5300  & 0.383 \tabularnewline
		\cline{2-5} \cline{3-5} \cline{4-5} \cline{5-5} 
		& DQN Multi-agent  & 00:51:14  & 8600  & 0.095 \tabularnewline
		\cline{2-5} \cline{3-5} \cline{4-5} \cline{5-5} 
		& PPO Centralized  & 00:21:26  & 600  & 0.555 \tabularnewline
		\cline{2-5} \cline{3-5} \cline{4-5} \cline{5-5} 
		& PPO Multi-agent  & \textbf{00:11:14}  & \textbf{800}  & \textbf{0.36} \tabularnewline
		\cline{2-5} \cline{3-5} \cline{4-5} \cline{5-5} 
		& Optimal Control  & -  & -  & 3.976 \tabularnewline
		\hline 
		\multirow{5}{2.6cm}{15 Workstations, 10 Tasks} & DQN Centralized  & 03:30:48  & 6100  & 0.698 \tabularnewline
		\cline{2-5} \cline{3-5} \cline{4-5} \cline{5-5} 
		& DQN Multi-agent  & 02:45:05  & 9400  & 0.095 \tabularnewline
		\cline{2-5} \cline{3-5} \cline{4-5} \cline{5-5} 
		& PPO Centralized  & 01:45:00  & 740  & 0.58 \tabularnewline
		\cline{2-5} \cline{3-5} \cline{4-5} \cline{5-5} 
		& PPO Multi-agent  & \textbf{00:25:29}  & \textbf{1050}  & \textbf{0.42} \tabularnewline
		\cline{2-5} \cline{3-5} \cline{4-5} \cline{5-5} 
		& Optimal Control  & -  & -  & 45.216 \tabularnewline
		\hline 
		\multirow{5}{2.6cm}{10 Workstations, 15 Tasks} & DQN Centralized  & 03:05:48  & 5390  & 0.703 \tabularnewline
		\cline{2-5} \cline{3-5} \cline{4-5} \cline{5-5} 
		& DQN Multi-agent  & 02:48:56  & 9600  & 0.095 \tabularnewline
		\cline{2-5} \cline{3-5} \cline{4-5} \cline{5-5} 
		& PPO Centralized  & 01:19:26  & 760  & 0.555 \tabularnewline
		\cline{2-5} \cline{3-5} \cline{4-5} \cline{5-5} 
		& PPO Multi-agent  & \textbf{00:19:14}  & \textbf{820}  & \textbf{0.36} \tabularnewline
		\cline{2-5} \cline{3-5} \cline{4-5} \cline{5-5} 
		& Optimal Control  & -  & -  & 45.436 \tabularnewline
		\hline 
	\end{tabular}
\end{table*}

Table \ref{tab5} presents the convergence episodes and training times for all the DRL algorithms, both centralized and decentralized. Additionally, it compares the inference time (i.e., the time required to compute a solution) with the time needed to solve the same problem using the optimal control (OC) method presented in \cite{10174426}. The solutions provided by the proposed DRL algorithms are more than 10 times faster than those obtained using the OC method, even in a small case study involving 3 workstations and 5 tasks.
%
%
%
%
%
%
%
%
%

\section{Conclusion} \label{sec:Conclusion}
This paper presented a mathematical model for a generalized industrial assembly line, based on the Markov Decision Process (MDP), which does not assume a specific assembly line type, in contrast to most existing models. This model was utilized to construct a virtual environment in which Deep Reinforcement Learning (DRL) agents are trained to optimize task and resource scheduling. Additionally, the paper introduces strategies to reduce agent training time. First, an action masking technique was implemented to ensure agents only apply feasible actions, significantly reducing training duration. Second, a multi-agent approach was developed, assigning each workstation its own agent, which promotes inter-agent cooperation and minimizes overall task completion time. This proposed multi-agent architecture offers a  learning framework for optimizing industrial assembly lines, enabling agents to learn offline and subsequently deliver real-time solutions by mapping the factory’s current state to optimal actions via a neural network, thus minimizing task duration. A comparative study of two Reinforcement Learning (RL) algorithms was conducted, with the introduction of action masking to improve learning efficiency and reduce training time. Moreover, the problem was decentralized into a Multi-Agent Reinforcement Learning (MARL) framework, reducing the action space’s growth rate from exponential to polynomial. A sequential training algorithm was developed to ensure that selected actions remain feasible. Numerical simulations demonstrate that the proposed multi-agent framework reduces training time by nearly tenfold compared to a centralized framework. The Proximal Policy Optimization (PPO) algorithm exhibited superior performance, converging to optimal solutions more effectively than the traditional Deep Q-Learning algorithm. Future work will address the potential discrepancies between the simulated environment used for training the DRL agent and the actual deployment environment.

\section*{Acknowledgment}

This work was supported in part by the National Sciences and Engineering
Research Council of Canada (NSERC) under the grants RGPIN-2022-04937.

\section*{CRediT authorship contribution statement}
Ali Mohamed Ali: Writing the original draft, Investigation, Methodology, Software, Methodology, Visualization, Formal
analysis.
Luca Tirel: Writing the original draft, Investigation, Methodology, Conceptualization, Software, Methodology, Visualization, Formal
analysis.
Hashim A. Hashim: Writing review \& editing, Conceptualization, Visualization, Supervision, Investigation, Validation, Funding acquisition.

\section*{Data availability}
Data will be made available on request.

\section*{Declaration of competing interest}
The authors declare that they have no known competing financial interests or personal relationships that could have appeared to influence the work reported in this paper.

%
%
\appendix
\section{Constraints Formulation in Reinforcement Learning Framework}
The constraints used to generate the mask are defined as follows, where the round brackets denote specific elements. We referred to $z \in [ 1, \dots, \delta_A ]$ as the index of a specific action in the actions space $A$, and $m(z)$ as the corresponding value:

\begin{enumerate}
	\item Do not assign finished tasks:
	\begin{equation} \label{eq:RLconst34}
		\begin{aligned} 
			if \: a[k](i,j) = 1 \And f[k](j) = 1  \longrightarrow m(z) = 0.
		\end{aligned}
	\end{equation}
	\item Do not assign a task if it is already in an execution state:
	\begin{equation} \label{eq:RLconst2}
		if \: a[k](i,j) = 1 \And \sum_{i \in |I|} e[k](i,j) = 1  \longrightarrow m(z) = 0.
	\end{equation}
	\item  Check deadlines are not violated when assigning a task: 
	\begin{gather*} \label{eq:const11}
		\begin{aligned} 
			\begin{gathered}
				if \: a[k](i,j) = 1 \\
				\And k + D(i,j) > F(j)  \longrightarrow m(z) = 0.
			\end{gathered}
		\end{aligned}
	\end{gather*}

	\item Check the current workload for each workstation does not exceed maximum occupancy:
	\begin{equation} \label{eq:const1}
		\begin{aligned} 
			\forall i, \: if \: o[k](i) + \sum_{j \in |J|} a[k](i,j) > O(i) \longrightarrow m(z) = 0.
		\end{aligned}
	\end{equation}
	
	\item Check that the time precedences are not violated when assigning tasks:
	\label{eq:const12}
	\begin{equation}
		\begin{gathered}
			if \: a[k](i,j_1) = 1 \\ \forall \: j_{2} \in P(j_1,j_2), \: if \: j_{2}=-1 \And \: f[k](j_2) = 0 
			\\ 
			\longrightarrow m(z) = 0.
		\end{gathered}
	\end{equation}
	
	\item Resources required must not exceed the available space in workstations’ buffers: 
	\begin{equation} \label{eq:const13}
		\begin{gathered}
			\forall i, \forall r, \: if \: y[k](i,j,r) = 1
			\\ 
			if \sum_{j \in |J|} C(j,r) + \sum_{j \in |J|} y[k](i,j,r) > U(i,r)
			\\ 
			\longrightarrow m(z) = 0.
		\end{gathered}
	\end{equation}

	\item Resources needed must exist in the factory:
	\begin{equation} \label{eq:RLconst42}
		\begin{gathered} 
			\forall r, \: if \: \sum_{i \in |I|} \sum_{j \in |J|} \sum_{r \in |R|} y[k](i,j,r)*C(j,r) > g[k](r)
			\\ 
			\longrightarrow m(z) = 0.
		\end{gathered}
	\end{equation}
\end{enumerate}

\balance
\bibliographystyle{IEEEtran}
\bibliography{ELS_Factory}
		
	\end{document}